\documentclass[letterpaper, 10 pt, conference]{ieeeconf}  % Comment this line out if you need a4paper

\IEEEoverridecommandlockouts                              % This command is only needed if
\usepackage{amssymb}
\usepackage{amsmath}
\usepackage{booktabs}
\usepackage{float}
\usepackage{graphicx}
\usepackage{multirow}
\usepackage{algorithm}
\usepackage{algpseudocode}
\usepackage{tabularx}
\let\labelindent\relax
\usepackage{enumitem}
\usepackage{cite}
\usepackage{hyperref}
\title{\LARGE \bf
Humanoid-DART: \underline{Humanoid} Loco-Manipulation using \underline{D}iffusion-guided \underline{A}ugmentation through \underline{R}elabeling and \underline{T}racking
}

\author{
\authorblockN{
Pranav Debbad, Kanish Thiagarajan, Victor Dhédin, Shafeef Omar, Majid Khadiv
}
\authorblockA{
Munich Institute of Robotics and Machine Intelligence
(MIRMI), Technical University of Munich (TUM), Germany.\\
Email: firstname.lastname@tum.de
}
}

\newcommand{\papertitle}{Humanoid-DART}

\begin{document}

\maketitle
\thispagestyle{empty}
\pagestyle{empty}

%%%%%%%%%%%%%%%%%%%%%%%%%%%%%%%%%%%%%%%%%%%%%%%%%%%%%%%%%%%%%%%%%%%%%%%%%%%%%%%%
\begin{abstract}

Imitating human demonstrations has emerged as a dominant paradigm for learning humanoid loco-manipulation policies. However, scaling these approaches remains challenging due to the high cost of collecting diverse demonstrations and the need for continual human intervention to correct policy failures. In this paper, we present a self-supervised framework that bootstraps from sparse demonstrations and progressively expands its behavioral repertoire, enabling the learning of a goal-conditioned policy that automatically explores the goal space with minimal expert supervision. Our approach combines diffusion-based trajectory generation with reinforcement learning, where the latter is used to track goal-conditioned trajectories produced by the diffusion model for a range of loco-manipulation skills. Through extensive ablation studies and comparisons with state-of-the-art methods, we demonstrate the effectiveness of our framework on multiple humanoid loco-manipulation skills. A summary of the results can be found  \href{https://www.youtube.com/watch?v=ORAL3zjS2O0}{here}.

\end{abstract}

%%%%%%%%%%%%%%%%%%%%%%%%%%%%%%%%%%%%%%%%%%%%%%%%%%%%%%%%%%%%%%%%%%%%%%%%%%%%%%%%
\section{Introduction}
Humanoid loco-manipulation requires coordinated whole-body control together with dynamic object interactions. Recent advances in imitation learning and reinforcement learning (RL) have demonstrated impressive humanoid skills by tracking expert demonstrations \cite{
he2024learning, liao2025beyondmimicmotiontrackingversatile}. In these approaches, demonstrations are typically collected through motion capture or teleoperation, retargeted to the robot embodiment, and used to train tracking-based RL policies. While recent generalist humanoid tracking policies \cite{yin2025unitrackerlearninguniversalwholebody, luo2026sonicsupersizingmotiontracking} achieve broad locomotion capabilities, their success relies heavily on the availability of large-scale human motion datasets \cite{AMASS:ICCV:2019}.

Extending this paradigm to contact-rich loco-manipulation remains substantially more difficult. Unlike locomotion, loco-manipulation involves large continuous task spaces defined jointly by body motion, object states, and environmental interactions. Variations in object pose, transport configuration, grasp strategy, and interaction timing create a combinatorial space that is prohibitively expensive to cover with demonstrations alone. Consequently, despite progress in general motion tracking, learning a single policy capable of robustly solving diverse loco-manipulation tasks under varying environmental conditions remains an open challenge, with existing methods typically focusing on tracking a single reference behavior \cite{yang2025omniretargetinteractionpreservingdatageneration, dhedin2026dynaretargetdynamicallyfeasibleretargetingusing}.

\begin{figure}[t]
    \centering
    \includegraphics[width=\linewidth]{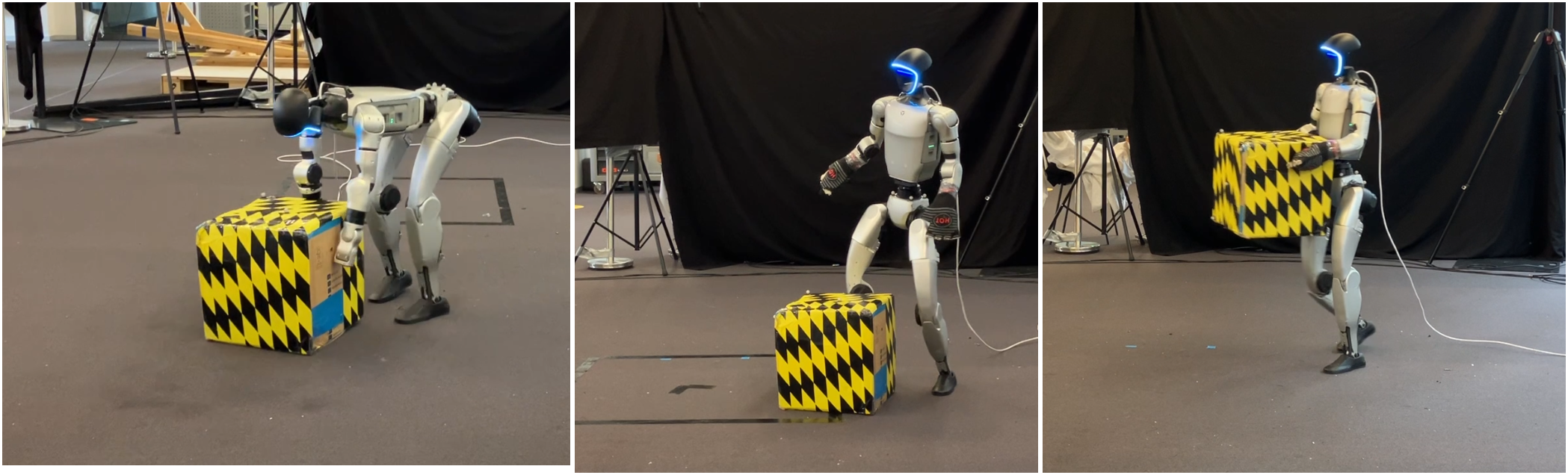}
    \caption{Real-world deployment: {\papertitle} trajectories deployed on a physical Unitree G1 humanoid for the \textit{push}, \textit{kick}, and \textit{pick-and-place} tasks (left to right).}
    \label{fig:hardware}
\end{figure}

In this work, we propose a curriculum-like framework that progressively expands the distribution of feasible goal-conditioned behaviors from only a small set of initial demonstrations. Our pipeline alternates between training a goal-conditioned motion generator and a motion-tracking policy, allowing both components to bootstrap one another toward increasingly diverse goals. Compared to policy-only RL curricula, this decomposition provides several advantages: the motion generator enables structured exploration in trajectory space and improves behavioral diversity in sparse-data settings; tracking rewards provide dense, task-agnostic supervision without extensive reward engineering; and separating high-level motion generation from low-level control reduces the burden on the generative model, enabling a more efficient and scalable learning pipeline.

In summary, this paper makes the following contributions:

\begin{itemize}[noitemsep]
    \item We present {\papertitle}, to the best of our knowledge, the first iterative trajectory augmentation pipeline for learning
    goal-conditioned loco-manipulation skills.
    \item We introduce three key design choices. A hierarchical goal-conditioned Diffusion Transformer (DiT) for generating humanoid \textit{and} object reference trajectories; an RL whole-body controller, conditioned on the object pose, to track those trajectories; and a curriculum-based algorithm with a goal relabeling strategy that converts near-missed rollouts into valid training signals while guiding exploration across the task space.
    \item We introduce two architecture choices for the motion generator that improve novel trajectory synthesis in the sparse-data regime: a \textit{dual-branch} diffusion transformer architecture that separates root and object motion from local features such as joints, and a \textit{structured partial unmasking} scheme that selectively reveals future feature groups during training to learn useful correlations.
    \item We validate {\papertitle} on a suite of humanoid loco-manipulation tasks (push, kick, hand-off, and pick \& place), demonstrating task-space coverage from as few as four base demonstrations and generalization to goals well
    beyond the seed distribution.
\end{itemize}

%%%%%%%%%%%%%%%%%%%%%%%%%%%%%%%%%%%%%%%%%%%%%%%%%%%%%%%%%%%%%%%%%%%%%%%%%%%%%%%%

\section{Related Work}
\label{sec:relatedwork}

\subsection{Loco-Manipulation from Human Demonstration}

% The high-dimensional and underactuated dynamics of humanoid robots make loco-manipulation
% especially difficult.
A dominant paradigm to generate humanoid loco-manipulation behaviors is to collect demonstrations via teleoperation and train imitation-learning policies from the resulting data \cite{seo2023deep}.
To reduce the cost of direct data collection, retargeting pipelines transfer human motions to robot embodiments: \cite{he2024learning} retargets kinematic pose sequences, while \cite{yang2025omniretargetinteractionpreservingdatageneration} explicitly preserves object and terrain contact relationships through interaction meshes. \cite{dhedin2026dynaretargetdynamicallyfeasibleretargetingusing} refines kinematic retargeted trajectories (which usually contain artifacts) into dynamically feasible ones which improved the RL training efficiency and performance.
Despite these advances, these approaches remain fundamentally limited by the coverage of the initial dataset. {\papertitle} addresses this bottleneck by leveraging generative models to iteratively expand the set of trajectory references, achieving an increasingly larger set of goals.

\subsection{Diffusion Models for Human/Humanoid Motion Synthesis}

Diffusion models~\cite{ho2020denoisingdiffusionprobabilisticmodels} have become a powerful class of generative models for high-dimensional synthesis such as image generation. They are used in motion generation~\cite{tevet2022humanmotiondiffusionmodel} and character control~\cite{Huang_2025} to synthesize diverse motions from text or action labels. ~\cite{kalaria2025dreamcontrolhumaninspiredwholebodyhumanoid, liao2025beyondmimicmotiontrackingversatile} extend these works to humanoid robots for tracking motion sequences. \cite{Kimodo2026} proposes a two-stage denoiser that decomposes root and body motion to suppress artifacts, which we adapt to our dual-branch transformer architecture. However, these works depend on the availability of large-scale data, while {\papertitle} aims to augment a small initial dataset containing sparse demonstrations, leveraging diffusion motion generation for kinematic trajectory exploration rather than end-to-end control. Additionally, we discuss some design decisions and training techniques for the diffusion model that enhance its expressiveness, enabling it to plan higher quality motions for novel goals.

\subsection{Curriculum Learning and motion synthesis}

In RL, curriculum learning refers to training agents by adapting task distributions or success criteria so that a single policy gradually acquires more advanced behaviors, leveraging what it previously learned \cite{JMLR:v21:20-212}. This is common practice in humanoid control as well as character animation \cite{huang2025learninghumanoidstandingupcontrol, zhang2024wococo, LeeLearning2021} but usually rely on a single policy that must both explore and acquire skills through environment interaction, often requiring careful reward design to cover a broad range of behaviors.

Related to our work, PARC \cite{xu2025parc} introduces an iterative simulation-based augmentation framework that alternates between motion generation and physics-based tracking. Humanoid-DART addresses a fundamentally different problem: goal-conditioned loco-manipulation over a continuous object-goal space, where task completion requires explicit reasoning over robot-object contact. Beyond architecture, we introduce a curriculum and goal relabeling scheme to achieve coverage in a continuous object-goal space from as few as four seed demonstrations. Furthermore, we go beyond character animation and show that the new goals can be achieved on a real humanoid robot.

%%%%%%%%%%%%%%%%%%%%%%%%%%%%%%%%%%%%%%%%%%%%%%%%%%%%%%%%%%%%%%%%%%%%%%%%%%%%%%%%

\section{Methodology}
\label{sec:methodology}
\begin{figure*}[t]
    \centering
    \includegraphics[width=0.95\linewidth]{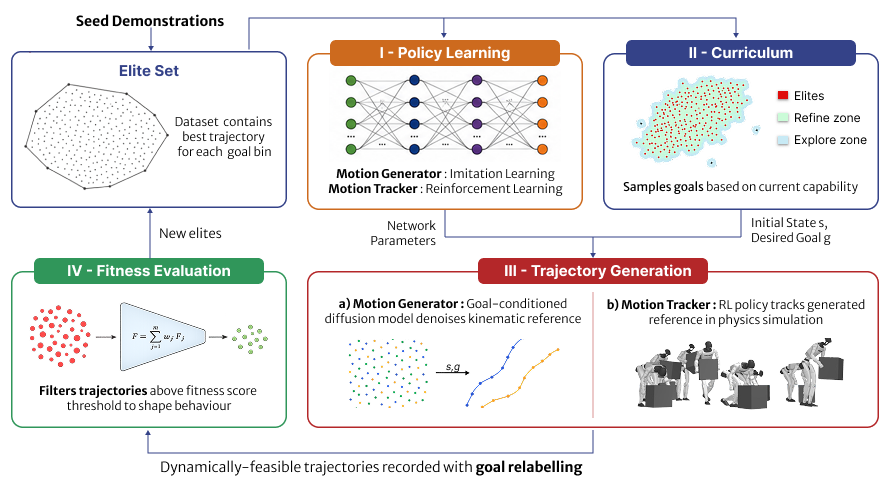}
    \caption{Overview of the proposed {\papertitle} pipeline.}
    \label{fig:overview}
\end{figure*}

In this work, we present {\papertitle}, a pipeline that learns loco-manipulation skills from sparse demonstrations. An overview of the pipeline can be seen in Fig.~\ref{fig:overview}.
Our framework consists of three tightly coupled components: a \emph{Kinematic Motion Generator} implemented as a task-conditioned diffusion model, a \emph{Physics-Based Evaluator} that validates generated motions in simulation, discriminating physically feasible motions from infeasible ones, and a \emph{Motion Tracking Policy} trained to execute the generated trajectories.
% We later provide a high-level overview of each component (Sections \ref{sec:kinematic_motion_generator}, \ref{sec:physics_based_motion_evaluator}, \ref{sec:dynamic_motion_tracker} respectively), with details in the Appendix. 
The rest of this section focuses on the main algorithm (Section \ref{sec:algo}) and the curriculum (Section \ref{sec:curriculum}), which are key novelties of the proposed pipeline.

\subsection{Algorithm}
\label{sec:algo}

The pipeline operates as an iterative curriculum over the current skill distribution, defined by the set of task goals covered by the existing trajectory archive $\mathcal{E}$. At each iteration, we sample $N_{\text{sampled}}$ goal states around this distribution, effectively expanding exploration to nearby but previously underrepresented regions of the task space. Conditioned on these goals, the motion generator $\pi_\theta$ produces kinematic reference trajectories, which may contain physical inconsistencies such as foot sliding, penetrations, or dynamic infeasibility.
These kinematic references are then executed in a physics simulator using the motion tracking policy $\pi_\phi$, which acts as a dynamic filter by converting potentially infeasible motion plans into physically consistent rollouts. The resulting trajectories are evaluated using a fitness function $F(\tau)$, and only those exceeding a threshold $\tau_f$ are retained in the elite archive $\mathcal{E}$. Additionally, before being added to the archive, rollouts are relabelled to preserve valid goal-conditioned demonstrations.
Finally, both components are updated using the expanded archive: the motion generator is retrained on $\mathcal{E}$ to improve goal-conditioned synthesis, while the tracking policy is refined via reinforcement learning to better reproduce the growing set of behaviors. Repeating this cycle progressively expands both the diversity of feasible trajectories and the coverage of the task space, enabling learning from a minimal set of initial demonstrations in a sparse-data regime (see Algorithm \ref{alg:dart}).

\begin{algorithm}[t]
\caption{\papertitle}
\label{alg:dart}
\begin{algorithmic}[1]

\Require Base demonstrations $\mathcal{D}_0$, fitness threshold $\tau_f$,
         sampling budget $N_{\text{sampled}}$
\Ensure  Elite archive $\mathcal{E}$, diffusion model $\pi_\theta$,
         motion tracker $\pi_\phi$

\If{$\mathcal{D}_0$ is kinematically feasible (KF)}
    \State Pretrain $\pi_\phi$ on $\mathcal{D}_0$; roll out $\mathcal{D}_0$ under $\pi_\phi$
           \Comment{dynamic feasibility bootstrapping}
    \State $\mathcal{E} \leftarrow \{\tau^* \mid F(\tau^*) \geq \tau_f\}$
           \Comment{seed archive with DF rollouts}
\Else
    \State $\mathcal{E} \leftarrow \mathcal{D}_0$
           \Comment{DF demonstrations seed archive directly}
\EndIf
\State Pretrain $\pi_\theta$ on $\mathcal{E}$
       \Comment{Motion generator pretraining}

\For{each generation $g$}
    \For{each iteration $k$}
        \State Sample $\{g_i\}_{i=1}^{N_{\text{sampled}}}$ via curriculum over $\mathcal{E}$
               \Comment{Goal sampling}
        \State Generate kinematic trajectories
               $\hat{\tau}_i \sim p_\theta(\tau \mid \tau_{\text{hist}}, g_i)$
               \Comment{Goal-conditioned diffusion}
        \State Roll out $\hat{\tau}_i$ under $\pi_\phi$;
               compute fitness $F_i$ $\forall$ $\tau^*_i$
               \Comment{Physics evaluation}
        \State $\mathcal{E} \leftarrow \mathcal{E} \cup \{\tau^*_i \mid F_i \geq \tau_f\}$
               \Comment{Update elites with goal relabeling}
        \State Retrain $\pi_\theta$ on updated $\mathcal{E}$
               \Comment{Motion Generator Training}
    \EndFor
    \State Retrain $\pi_\phi$ on $\mathcal{E}$ via PPO
           \Comment{Motion Tracker training}
\EndFor

\State \Return $\mathcal{E},\, \pi_\theta,\, \pi_\phi$

\end{algorithmic}
\end{algorithm}

\subsection{Kinematic Motion Generator}\label{sec:kinematic_motion_generator}
For motion generation, we adopt a generative modeling approach, using a goal-conditioned diffusion transformer to produce motions that lie close to the demonstration manifold. The generated motions are thus kinematically plausible and are taken as reference trajectories for the downstream tracking controller. The key design decisions for improving the expressivity and the kinematic consistency of the motion generator were the motion representation, the network architecture, and training and inference techniques. Namely, we introduce a \textit{dual-branch backbone} and \textit{structured-feature unmasking} that increased the quality of the generated motion, as shown later in the results.

\paragraph{Motion Representation}
The expert demonstrations are converted into a segmented egocentric representation to increase data density and enable smoother motion stitching across trajectories~\cite{seo2023deep, tevet2022humanmotiondiffusionmodel}. All features are expressed relative to the robot's local coordinate frame at $t_0$, with trajectories translated and yaw-aligned~\cite{luo2024universalhumanoidmotionrepresentations} with respect to the robot's global heading at $t_0$. This normalization removes dependence on global position and heading, allowing motion priors from different demonstrations to be more easily combined. The robot and object state at each frame $t$ is represented by:
\begin{equation} \label{eq:vector_representation}
    \tau_t = [\Delta p_{rob,xy},\; p_{rob,z},\; \Delta \psi_{rob},\; j_{pos},\; r_{rob,xy},\; {}^{rob} p_{obj},\; {}^{rob} r_{obj}]
\end{equation}
where $\Delta p_{rob,xy}\in\mathbb{R}^2$ is the root x--y displacement in the yaw-aligned egocentric frame at $t_0$, $p_{rob,z}\in\mathbb{R}$ the absolute root height in the world frame, $\Delta\psi_{rob}\in\mathbb{R}$ the root yaw rotation about the $z$-axis, $j_{pos}\in\mathbb{R}^{N_j}$ the joint angles relative to the default pose, $r_{rob,xy}\in\mathbb{R}^6$ the root pitch and roll in the 6D rotation representation~\cite{zhou2019continuity}, and ${}^{rob}p_{obj}\in\mathbb{R}^3$, ${}^{rob}r_{obj}\in\mathbb{R}^6$ the object position and orientation (6D) in the robot base frame. Robot heading (yaw) is separated from the remaining base orientation to mirror the hierarchical global/local structure of the generator; the robot-relative object pose mitigates kinematic artifacts such as floating and penetration.

\paragraph{Network Architecture}
Two design choices distinguish our generator from a standard diffusion transformer.
\textbf{(i) Dual-branch backbone.}  Factorizing the state into world features (navigation) and local features (body kinematics) has been shown to improve motion diversity in human motion generation~\cite{Kimodo2026}. Here, the motion generator is a dual-branch DiT1D with a \textit{global stream} over root motion and object-relative features, conditioned on task goals and motion history via AdaLN~\cite{peebles2023dit} modulation, and a \textit{local stream} over body pose features. The local stream cross-attends to the global stream to stay synchronized with the global trajectory, decoupling coarse navigation from fine-grained pose synthesis but allowing conditioning.
\textbf{(ii) Structured partial unmasking.} Inspired by~\cite{chen2024diffusionforcingnexttokenprediction, tessler2024maskedmimicunifiedphysicsbasedcharacter}, which expose the denoiser to partial contexts, our motion generator uses per-token noise levels and selectively reveals noised feature groups from future frames during training in order to learn partial conditioning. This encourages the planner to infer the remaining motion components from partially observed states and to learn correlations between feature groups rather than treating each independently. Network dimensions and training details are provided in the Appendix.
% $\{\Delta p_{rob,xy}, \Delta\psi_{rob}, p_{rob,z}, {}^{rob}p_{obj}, {}^{rob}r_{obj}\}$
% $\{j_{pos}, r_{rob,xy}\}$

\paragraph{Inference}
At inference, the motion denoiser models the conditional distribution $p_{\theta}(\tau_{(t_0-t_h):(t_0+t_p)} \mid \tau_{(t_0-t_h):t_0},\, g)$ using $10$ deterministic DDIM steps. Long-horizon sequences are synthesised auto-regressively~\cite{lin2025simgenhoi} over overlapping segments, each conditioned on $h$ history frames from the previous segment and task goal $g$. History frames are inpainted (RePaint-style)~\cite{lugmayr2022repaint} during denoising to enforce temporal consistency, and classifier-free guidance~\cite{ho2022classifierfreediffusionguidance} is applied with a tunable guidance weight.

\begin{figure}[t]
    \centering
    \includegraphics[width=\linewidth]{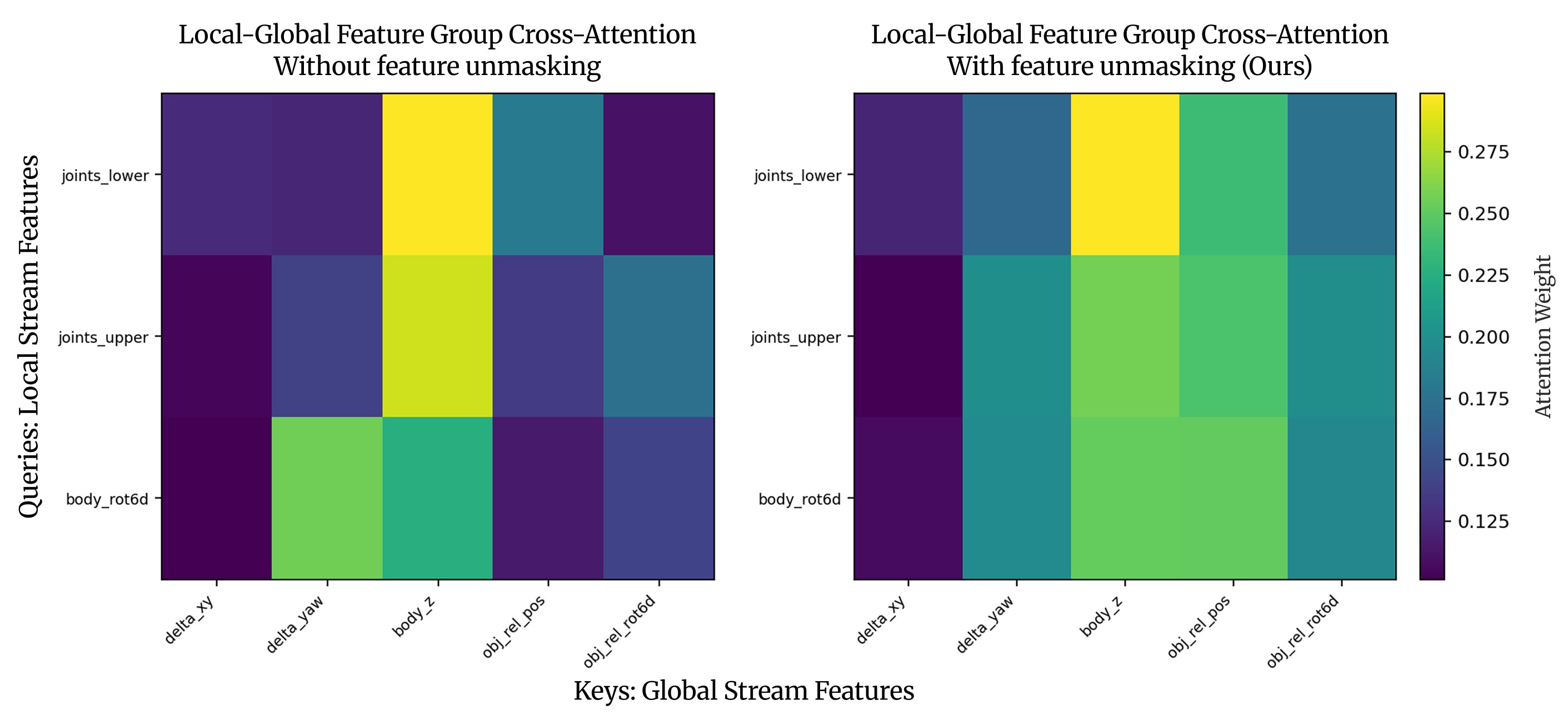}
    \caption{Effect of structured partial unmasking on local-to-global cross-attention. Structured partial unmasking increases the attention of robot joints and pose to the object pose.}
    \label{fig:cross_attn}
\end{figure}

\paragraph{Cross-attention analysis}
To verify that the local stream exploits the global stream as intended, we inspect the local-to-global cross-attention of the trained motion generator with and without structured partial unmasking. Each query token in the local stream---a single feature group of $\tau_t$ at one time step---attends to the global-stream key tokens; Figure~\ref{fig:cross_attn} reports the resulting group cross-attention from the local-stream queries (rows) to the global-stream keys (columns), averaged over the attention heads and cross-attention layers. Without unmasking (Fig.~\ref{fig:cross_attn}, Left), attention collapses heavily onto the root height $p_{rob,z}$ and largely ignores the object-relative pose. With unmasking (Right), the local stream additionally attends to the object-relative position and orientation (${}^{rob}p_{obj}, {}^{rob}r_{obj}$), recovering a correlation between whole-body pose synthesis and object interaction that is highly relevant to humanoid loco-manipulation.

\subsection{Physics-Based Motion Evaluator}\label{sec:physics_based_motion_evaluator}

In each iteration of the evolutionary pipeline, the motion generator produces $N_{\text{sampled}}$ candidate trajectories attempting to realize different tasks, of which several may be dynamically infeasible, untrackable, or fail to solve the task. A rigorous selection mechanism is therefore essential to keep only the high-quality \textit{elites}.
The fitness function evaluates physical feasibility and task correctness as the product of several terms:
\[
\text{Fitness}
=
R_{\text{pos}}
\cdot R_{\text{vel}}
\cdot R_{\text{contact}}
\cdot R_{\text{root,pos}}
\cdot R_{\text{root,ori}}
\cdot R_{\text{leg}}
\cdot R_{\text{obj}}.
\]

These terms evaluate the quality of the generated motions along two axes. \textit{Tracking fidelity} is measured via Gaussian kernel terms $\exp(-\|\cdot\|^2/\sigma^2)$ over key body positions ($R_{\text{pos}}$, $\sigma{=}0.05$), link velocities ($R_{\text{vel}}$, $\sigma{=}1.0$), pelvis-root position ($R_{\text{root,pos}}$, $\sigma{=}0.05$) and orientation ($R_{\text{root,ori}}$, $\sigma{=}0.4$), and lower-limb joint pose ($R_{\text{leg}}$, $\sigma{=}0.04$), each penalizing deviation from the reference trajectory. \textit{Contact correctness} combines an indicator term $R_{\text{contact}}$ that checks whether the expected hand--object contacts are made at each timestep with a Gaussian term $R_{\text{obj}}$ ($\sigma{=}1.0$) on the 6-DoF object pose error. A zero-gate sets the fitness to $0$ on a fall or early termination.

\subsection{Dynamic Motion Tracker}\label{sec:dynamic_motion_tracker}
We implement a DeepMimic-inspired \cite{peng2018deepmimic} tracking controller extended to handle object-robot contacts. Rewards penalise deviations from the reference root pose, body keypoints, joint positions, and end-effector contact states (full reward terms in the Appendix). The asymmetric actor-critic policy observes a 5-step reference horizon, proprioceptive state, and object pose; the critic additionally receives privileged simulation state. Following \cite{liao2025beyondmimicmotiontrackingversatile}, we use adaptive trajectory- and state-level sampling (episodes initialized mid-trajectory) to accelerate learning. Training uses massively parallelized Proximal Policy Optimization (PPO) with Generalized Advantage Estimation (GAE) and domain randomization over object pushes, friction, and mass for robust sim-to-real transfer. Network sizes, observation terms, PPO hyperparameters, and domain-randomization ranges are provided in the Appendix.

\subsection{Curriculum}\label{sec:curriculum}
The order in which goals in the task space are sampled affects the pipeline's learning dynamics, both the rate at which the task space is covered and the stability of convergence. We define a task-space curriculum that governs how targets are queried across iterations. New goals are sampled inside and around an elite set $\mathcal{E}$ to both refine and explore skills. We additionally relabel near goal successes to provide a better learning signal.

\paragraph{Frontier Target Sampling}
\label{sec:frontier}

Let \(\mathcal{B}\) denote the set of all task-space bins induced by the
task configuration and let \(c_b \in \mathbb{R}^{d}\) be the center of
bin \(b\). For example, in the pick-and-place task, the task space is the 2D goal displacement $(x, y)$ of the box, discretised into a uniform
grid of bins with resolution $\Delta = 0.02$\,m.
Given the current elite set

\(\mathcal{E} = \{e_1, \dots, e_n\}\), the curriculum computes the minimum distance from the elites $d(b)$, for each bin $b$, we have $d(b) = \min_{e \in \mathcal{E}} \lVert \tilde{c}_b - \tilde{m}_e \rVert_2 ,$
where \(\tilde{c}_b\) is the normalized bin center and
\(\tilde{m}_e\) is the normalized task metric of elite \(e\).
Two distance thresholds partition the task space into three zones:

\emph{Refine}: ($\mathcal{R} = \{ b \in \mathcal{B} \mid d(b) < \delta_{\mathrm{ref}} \}$), \emph{eXplore}: ($\mathcal{X}= \{ b \in \mathcal{B} \mid \delta_{\mathrm{ref}} \le d(b) < \delta_{\mathrm{exp}} \}$), and
\emph{Frontier}: ($\mathcal{F} = \{ b \in \mathcal{B} \mid d(b) \ge \delta_{\mathrm{exp}}$) regions. Bins near existing elites are treated as
refinement targets, bins at intermediate distance encourage controlled
coverage expansion, and far-away bins represent as-yet unsolved regions
of the task space.
The archive assigns each bin a sampling weight; the probability of sampling a bin is the ratio of its weight to the sum of weights across all bins, i.e., 
$P(b) = \frac{w_b}{\sum_{b' \in \mathcal{B}} w_{b'}}$, 
where $w_b$ is $w_{\mathrm{ref}},\: \forall \: b \in \mathcal{R}$; $w_{\mathrm{exp}},\: \forall \: b \in \mathcal{X}$; and $w_{\mathrm{frontier}},\: \forall \: b \in \mathcal{F}$.

In the reference implementation, exploration bins receive the highest
weight, refinement bins receive a smaller but non-zero weight, and
frontier bins receive a small probability mass so that the search
occasionally proposes more ambitious outlying tasks. The concrete threshold and weight values, along with the remaining evolution hyperparameters, are listed in the Appendix.

\paragraph{Goal Relabeling}
\label{sec:relabel}

The diffusion planner is conditioned on a sampled target, but the executed trajectory need not match it exactly. Rather than treating such deviations as failures, the pipeline relabels physically valid rollouts with their \emph{achieved} task metrics. This mechanism is analogous to~\cite{andrychowicz2017her}, which recovers learning signal from failed goal-conditioned RL trajectories by retroactively treating the achieved state as the intended goal, and is particularly effective in early pipeline iterations where the generator is undertrained and systematic goal misses are common.

%%%%%%%%%%%%%%%%%%%%%%%%%%%%%%%%%%%%%%%%%%%%%%%%%%%%%%%%%%%%%%%%%%%%%%%%%%%%%%%%

\section{Experimental Results}
\label{sec:result}
We evaluate {\papertitle} through a combination of real-world deployment and large-scale simulation. As shown in Fig.~\ref{fig:hardware}, the loco-manipulation trajectories produced by our pipeline are deployed on a physical Unitree G1 humanoid, confirming that the generated motions are dynamically feasible on hardware. Our experiments are designed to answer the following questions:
\textbf{1)} Can extremely sparse skills be scaled to a generalized task space using a curriculum-based approach?
\textbf{2)} Can out-of-distribution skills be learned through a combination of goal-conditioned motion generation and reinforcement learning?
\textbf{3)} How does the quality and number of seed trajectories affect the pipeline's coverage rate and convergence?

\subsection{Implementation}
We use reference motions from DynaRetarget~\cite{dhedin2026dynaretargetdynamicallyfeasibleretargetingusing} containing hundreds of dynamically feasible object-robot loco-manipulation trajectories.
We implement {\papertitle} in MuJoCo using mjlab \cite{Zakka_mjlab_A_Lightweight_2026} for GPU-accelerated RL. The motion tracking policy runs at 50Hz, while the motion generator generates full sequences of trajectories through autoregressive stitching with inpainting to ensure successive frames are continuous. We use a timestep $\Delta t=0.005$ in MuJoCo with a decimation of 4 for the controller, and consider the full collision model of the robot. All experiments are run on a single NVIDIA RTX 5090 GPU.

We evaluate our pipeline on four loco-manipulation tasks of increasing complexity: \textit{push}, \textit{kick}, \textit{hand-off}, and \textit{pick-and-place}. In pick-and-place, the robot
must grasp an object and transport it to a target location within a
planar task space parameterised by goal displacement $(x, y)$ relative to the robot's starting root pose. Push and kick share the same planar parameterisation but require distinct contact modes,
whole-body pushing without a grasp, and foot-contact kicking
respectively. Hand-off requires the robot to present an object at a target height and distance.
For each task, we initialise the pipeline with a small set of $2$ to $4$ sparse
base demonstrations, each capturing a distinct reference trajectory
to a different goal configuration. Experiments use the Unitree G1 humanoid with $29$ actuated degrees of freedom, manipulating a box whose friction and mass are randomized about their nominal values.

% \begin{table}[h]
% \centering
% \caption{Task parameterizations used in evaluation.}
% \label{tab:task_params}
% \renewcommand{\arraystretch}{1.2}
% \begin{tabular}{llcc}
% \toprule
% \textbf{Task} & \textbf{Task Axis} & \textbf{Range} & \textbf{\#Base} \\
% \midrule
% \multirow{2}{*}{Pick \& Place}
%   & Final box $x$ (m) & $[-3.0,\ 5.0]$ & \multirow{2}{*}{4} \\
%   & Final box $y$ (m) & $[\ 4.4,\ 1.6]$ & \\
% \midrule
% \multirow{2}{*}{Push}
%   & Final box $x$ (m) & $[-0.5,\ 2.5]$ & \multirow{2}{*}{4} \\
%   & Final box $y$ (m) & $[\ -1.8,\ 1.2]$ & \\
% \midrule
% \multirow{2}{*}{Kick}
%   & Final box $x$ (m) & $[-0.5,\ 2.5]$ & \multirow{2}{*}{4} \\
%   & Final box $y$ (m) & $[\ -1.8,\ 1.2]$ & \\
% \midrule
% \multirow{2}{*}{Hand-off}
%   & Hand-off height (m)   & $[0.7,\ 1.3]$ & \multirow{2}{*}{2} \\
%   & Hand-off distance (m) & $[0.25,\ 0.55]$ & \\
% \bottomrule
% \end{tabular}
% \end{table}

The pipeline runs for $g=4$ generations, with $k=10$ iterations per generation. At each iteration, $N_{sampled}=3000$ candidate trajectories are sampled via the curriculum and the motion generator is updated. At the end of each generation, the motion tracker is retrained on the elite buffer.

\begin{figure*}[t]
    \centering
    \includegraphics[width=\linewidth]{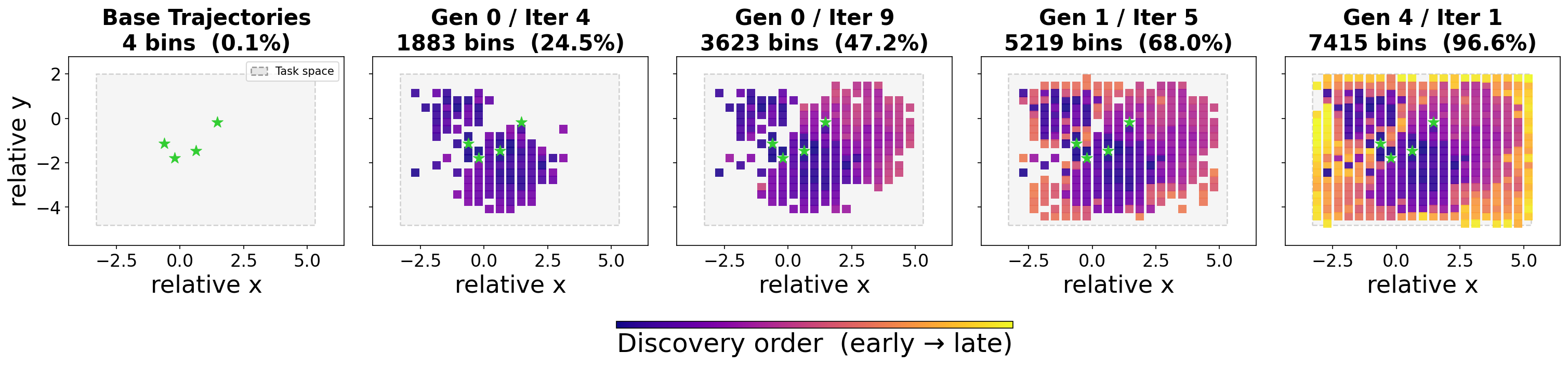}
    \caption{Task-space coverage over the pipeline. Bins coloured by discovery order (purple=earliest, yellow=latest); green stars are seed demos; dashed region is the target space. Coverage grows from 0.1\% to 96.6\%.}
    \label{fig:elites_progression}
\end{figure*}

\subsection{Curriculum Performance}

\begin{figure*}[t]
    \centering
    \includegraphics[width=\linewidth]{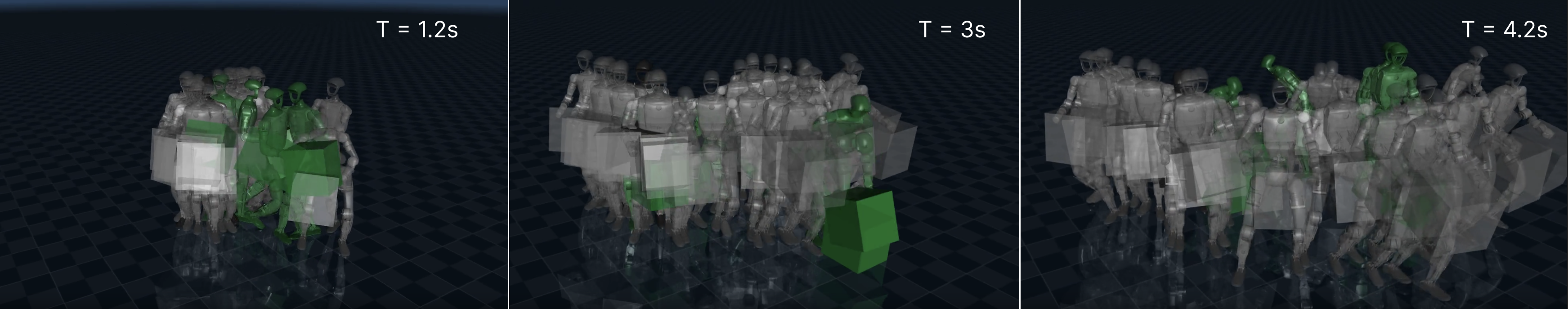}
    \caption{Family of learnt loco-manipulation skills for \textit{pick-and-place}. Green motions are the seed base trajectories; the others are generated by the \protect\papertitle{} motion generator.}
    \label{fig:pipeline_output}
\end{figure*}

Figure \ref{fig:elites_progression} illustrates the progression of elite trajectories at representative stages throughout the pipeline for the \textit{pick-and-place} task. In early iterations, generated trajectories cluster around the base demonstrations, as expected. As the evolutionary process advances, the elite set progressively expands its coverage of the task space. By the fourth generation, the pipeline achieves near-complete coverage of the target task space.
A notable emergent property of the pipeline is its ability to generate and track long-horizon, task-conditioned trajectories well beyond the scope of the base demonstrations. The base trajectories are approximately 5 seconds in duration, with goal placement distances in the range of 1--2\,m. Despite this, the pipeline successfully generates trajectories to goals 4-5x further away, demonstrating generalisation across the full task space. Fig.~\ref{fig:pipeline_output} shows the pipeline's output at various timestamps.

\subsection{Comparison with Baselines}
We evaluate against two baselines chosen to isolate the contributions of our pipeline. Parameterised Motion~\cite{LeeLearning2021} represents the closest prior work in the space of parameterised skill generation from sparse demonstrations. Following the paradigm of \cite{lin2025simgenhoi}, Hierarchical Diffusion + RL is a method in which the diffusion trajectory generator and the RL tracker are trained jointly in a single stage on the seed trajectories only, without the evolutionary loop. This isolates the benefit of our pipeline from the architectural choice of combining RL with diffusion.

As summarised in Table~\ref{tab:coverage_tasks}, Humanoid-DART achieves dominant task-space coverage across all four tasks, demonstrating the effectiveness of the iterative pipeline. The fitness and object stability results reveal a trade-off: Parameterised Motion achieves higher fitness on hand-off and pick-and-place, as it concentrates its limited coverage on a narrow region of the task space where it can produce high-quality trajectories. In contrast, the lower average fitness of Humanoid-DART for these tasks reflects the challenge of maintaining trajectory quality across a far broader and more diverse set of goals. Hierarchical Diff. + RL consistently underperforms on both metrics, confirming that the evolutionary loop is the primary driver of performance.

\begin{table}[h]
\centering
\caption{Final coverage, average fitness, and object stability across the task suite.}
\small
\label{tab:coverage_tasks}
\setlength{\tabcolsep}{4pt}
\resizebox{\linewidth}{!}{%
\begin{tabular}{llcccc}
\toprule
\textbf{Method} & \textbf{Metric}
  & \textbf{Push}
  & \textbf{Kick}
  & \textbf{H-off}
  & \textbf{P\&P} \\
\midrule
\multirow{3}{*}{\textbf{\papertitle}}
  & Cov.\,$\uparrow$  & \textbf{61.1} & \textbf{54.8} & \textbf{51.5} & \textbf{96.4} \\
  & Fit.$^\dagger$\,$\uparrow$ & \textbf{3.28} & \textbf{19.8} & 3.2 & 4.9 \\
  & Stab.$^\ddagger$\,$\uparrow$ & 0.60 & \textbf{0.59} & 0.43 & \textbf{0.60} \\
\cmidrule(lr){1-6}
\multirow{3}{*}{Param.~\cite{LeeLearning2021}}
  & Cov.\,$\uparrow$  & 3.86 & 23.89   & 24.4 & 5.7 \\
  & Fit.$^\dagger$\,$\uparrow$ & 2.92 & 4.79 & \textbf{6.91} & \textbf{10.0} \\
  & Stab.$^\ddagger$\,$\uparrow$ & \textbf{0.62} & 0.52 & \textbf{0.51} & 0.52 \\
\cmidrule(lr){1-6}
\multirow{3}{*}{Hier. D+RL}
  & Cov.\,$\uparrow$  & 2.2  & 2.6  & 32.0 & 6.2 \\
  & Fit.$^\dagger$\,$\uparrow$ & 1.1 & 1.77 & 4.1 & 0.9 \\
  & Stab.$^\ddagger$\,$\uparrow$ & 0.53 & 0.38 & 0.38 & 0.37 \\
\bottomrule
\end{tabular}}

\vspace{2pt}
{\footnotesize Cov.\ = coverage (\%). $^\dagger$Avg.\ fitness $\times10^{-3}$. $^\ddagger$Obj.\ deviation from ref., exp.\ penalised; 1 = perfect.}
\end{table}

\subsection{Motion Generator Architecture}
To justify our two main design decisions in isolation, we train each variant on the same four base loco-manipulation motions and evaluate it on a sweep of 20 \emph{novel} goals. All metrics are computed directly on the generated kinematic trajectories, \emph{without} the downstream RL tracking policy, in order to isolate the quality of the planner's output independent of the controller, and capture goal-reaching accuracy, path quality, and grasp consistency. Table~\ref{tab:ablations} highlights that the \textit{dual-branch backbone} is decisive for goal-reaching: collapsing both streams into a single flat DiT drops the success rate from $1.00$ to $0.25$, reflecting poor coordination between locomotion, manipulation, and whole-body pose. \textit{Structured partial unmasking} improves path quality and grasp consistency: straighter, shorter object paths and lower hand--object distance. The two components together increase the attention between the robot whole-body pose and the relative object pose, which is a useful correlation in humanoid loco-manipulation, as shown in Fig.~\ref{fig:cross_attn}.

\begin{table}[h]
\centering
\caption{Motion generator ablations on 20 novel goals. G.Err.: object--goal distance (m); Succ.: fraction within 20\,cm; Strt.: straightness; Path: object path length (m); H--o: hand--object distance airborne (m).}
\label{tab:ablations}
\small
\setlength{\tabcolsep}{4pt}
\begin{tabular}{lccccc}
\toprule
Variant & G.Err.\,$\downarrow$ & Succ.\,$\uparrow$ & Strt.\,$\uparrow$ & Path\,$\downarrow$ & H--o\,$\downarrow$ \\
\midrule
\textbf{Dual-branch}        & \textbf{0.115} & \textbf{1.00} & \textbf{0.68} & \textbf{4.70} & \textbf{0.26} \\
Dual (no unmask.)  & 0.121 & 1.00 & 0.58 & 6.30 & 0.32 \\
Single-branch           & 0.316 & 0.25 & 0.34 & 13.96 & 0.42 \\
\bottomrule
\end{tabular}
\end{table}

\subsection{Pipeline Ablations}
We conduct ablation studies varying the number of base demonstrations provided at initialisation, allowing us to isolate the effect of initialisation sparsity on downstream task-space coverage.
Additionally, we ablate the nature of the base trajectory used to seed the pipeline by considering artifact-free \textit{dynamically feasible} (DF) trajectories and \textit{kinematically feasible} (KF) trajectories, produced via kinematic retargeting only, which may contain contact inconsistencies and penetration.
Results are summarised in Table~\ref{tab:coverage_ablation}.

\begin{table}[h]
\centering
\caption{Coverage vs.\ number and type of base demonstrations. $t_{x\%}$: time (h) to reach $x\%$ coverage; \textbf{--}: not reached.}
\label{tab:coverage_ablation}
\small
\begin{tabular}{llccccc}
\toprule
\textbf{Init.} & \textbf{\#}
  & \textbf{$t_{20}$} & \textbf{$t_{50}$}
  & \textbf{$t_{80}$} & \textbf{Cov.\%} & \textbf{Fit.} \\
\midrule
\multirow{3}{*}{DF} & 1 & 1.82 & 3.64 & --   & 77.3 & $3.69$ \\
                    & 2 & 1.2  & 2.0  & 3.65 & 91.2 & $4.17$ \\
                    & 4 & 0.6  & 1.5  & 3.3  & 96.4 & $4.9$ \\
\cmidrule(lr){1-7}
KF                  & 2 & 2.38 & --   & --   & 28.8 & $1.37$ \\
\bottomrule
\addlinespace[4pt]
\multicolumn{7}{l}{\small Pick \& place task. Fitness scaled by $\times10^{-3}$.} \\
\end{tabular}
\end{table}

The gap between KF and DF initialization shows that the diffusion generator inherits and amplifies biases in the seed trajectories. Kinematically retargeted seeds with physical inconsistencies produce lower-quality candidates and slower coverage growth, whereas dynamically feasible and diverse seeds lead to higher-quality elites, faster convergence, and improved coverage.

%%%%%%%%%%%%%%%%%%%%%%%%%%%%%%%%%%%%%%%%%%%%%%%%%%%%%%%%%%%%%%%%%%%%%%%%%%%%%%%%

\section{Conclusions and future work}
\label{sec:conclusion}

We present {\papertitle}, an iterative pipeline for learning goal-conditioned humanoid loco-manipulation policies from sparse demonstrations. By coupling a goal-conditioned diffusion motion generator with a physics-based RL tracking controller through an evolutionary pipeline, {\papertitle} progressively expands the set of physically feasible behaviors across a continuous task space without large-scale human supervision. Across multiple loco-manipulation settings, and in particular pick-and-place, {\papertitle} achieves near-complete task-space coverage from as few as four base demonstrations (20 seconds of motion), generalizing to goals four to five times beyond the seed range.

Our evaluation is currently restricted to a single object geometry and flat terrain; extending the framework to more diverse objects, contact properties, and uneven terrain remains the focus of future work. The pipeline also inherits the biases of the seed demonstrations and depends on a carefully calibrated fitness function, which can be replaced with an adaptive, learned criterion that scores motion quality directly from data and adjusts its selectivity as coverage grows, reducing the reliance on hand-tuned terms.
% \section{Limitations}
% The pipeline inherits the stylistic and kinematic biases of the seed demonstrations, so artifacts such as unnatural contacts or poor posture can propagate and amplify during generation; the fitness function is therefore critical, as poorly calibrated thresholds can admit degenerate behaviors or overly restrict exploration. The generator is also limited by the diversity of the elite archive and cannot discover strategies outside the demonstrated manifold, so performance depends on the geometric diversity of the seed demonstrations, which still requires human curation. 

%%%%%%%%%%%%%%%%%%%%%%%%%%%%%%%%%%%%%%%%%%%%%%%%%%%%%%%%%%%%%%%%%%%%%%%%%%%%%%%%

\bibliographystyle{ieeetr}
\bibliography{references}

@misc{yang2025omniretargetinteractionpreservingdatageneration,
title={OmniRetarget: Interaction-Preserving Data Generation for Humanoid Whole-Body Loco-Manipulation and Scene Interaction}, 
author={Lujie Yang and Xiaoyu Huang and Zhen Wu and Angjoo Kanazawa and Pieter Abbeel and Carmelo Sferrazza and C. Karen Liu and Rocky Duan and Guanya Shi},
year={2025},
eprint={2509.26633},
archivePrefix={arXiv},
primaryClass={cs.RO},
url={https://arxiv.org/abs/2509.26633},
}

@inproceedings{zhou2019continuity,
  title={On the Continuity of Rotation Representations in Neural Networks},
  author={Zhou, Yi and Barnes, Connelly and Lu, Jingwan and Yang, Jimei and Li, Hao},
  booktitle={Proceedings of the IEEE/CVF Conference on Computer Vision and Pattern Recognition (CVPR)},
  pages={5745--5753},
  year={2019}
}

@article{LeeLearning2021,
author = {Lee, Seyoung and Lee, Sunmin and Lee, Yongwoo and Lee, Jehee},
title = {Learning a family of motor skills from a single motion clip},
year = {2021},
issue_date = {August 2021},
publisher = {Association for Computing Machinery},
address = {New York, NY, USA},
volume = {40},
number = {4},
issn = {0730-0301},
url = {https://doi.org/10.1145/3450626.3459774},
doi = {10.1145/3450626.3459774},
journal = {ACM Trans. Graph.},
month = jul,
articleno = {93},
numpages = {13}
}

@misc{dhedin2026dynaretargetdynamicallyfeasibleretargetingusing,
      title={DynaRetarget: Dynamically-Feasible Retargeting using Sampling-Based Trajectory Optimization}, 
      author={Victor Dhedin and Ilyass Taouil and Shafeef Omar and Dian Yu and Kun Tao and Angela Dai and Majid Khadiv},
      year={2026},
      eprint={2602.06827},
      archivePrefix={arXiv},
      primaryClass={cs.RO},
      url={https://arxiv.org/abs/2602.06827}, 
}

@inproceedings{xu2025parc,
author = {Xu, Michael and Shi, Yi and Yin, KangKang and Peng, Xue Bin},
title = {PARC: Physics-based Augmentation with Reinforcement Learning for Character Controllers},
year = {2025},
isbn = {9798400715402},
publisher = {Association for Computing Machinery},
address = {New York, NY, USA},
url = {https://doi.org/10.1145/3721238.3730616},
doi = {10.1145/3721238.3730616},
booktitle = {Proceedings of the Special Interest Group on Computer Graphics and Interactive Techniques Conference Conference Papers},
articleno = {131},
numpages = {11},
location = {
},
series = {SIGGRAPH Conference Papers '25}
}

@inproceedings{seo2023deep,
  title     = {Deep Imitation Learning for Humanoid Loco-Manipulation
               through Human Teleoperation},
  author    = {Seo, Mingyo and Han, Seungyeon and Sim, Kyungmin and
               Bang, Seung Hyeon and Gonzalez, Carlos and
               Sentis, Luis and Zhu, Yuke},
  booktitle = {IEEE-RAS International Conference on Humanoid Robots},
  year      = {2023}
}

@article{zhang2024wococo,
  title   = {{WoCoCo}: Learning Whole-Body Humanoid Control
             with Sequential Contacts},
  author  = {Zhang, Chong and Xiao, Wenli and He, Tairan and Shi, Guanya},
  journal = {arXiv preprint arXiv:2406.06005},
  year    = {2024}
}

@article{he2024learning,
  title   = {Learning Human-to-Humanoid Real-Time Whole-Body Teleoperation},
  author  = {He, Tairan and Luo, Zhengyi and Xiao, Wenli and
             Zhang, Chong and Kitani, Kris and Liu, Changliu and Shi, Guanya},
  journal = {arXiv preprint arXiv:2403.04436},
  booktitle = {IEEE/RSJ International Conference on Intelligent
               Robots and Systems (IROS)},
  year    = {2024}
}

@inproceedings{peng2018deepmimic,
  title     = {{DeepMimic}: Example-Guided Deep Reinforcement Learning
               of Physics-Based Character Skills},
  author    = {Peng, Xue Bin and Abbeel, Pieter and Levine, Sergey
               and van de Panne, Michiel},
  booktitle = {ACM Transactions on Graphics (SIGGRAPH)},
  volume    = {37},
  number    = {4},
  year      = {2018}
}

@inproceedings{andrychowicz2017her,
  title     = {Hindsight Experience Replay},
  author    = {Andrychowicz, Marcin and Wolski, Filip and Ray, Alex and
               Schneider, Jonas and Fong, Rachel and Welinder, Peter and
               McGrew, Bob and Tobin, Josh and Abbeel, Pieter and
               Zaremba, Wojciech},
  booktitle = {Advances in Neural Information Processing Systems (NeurIPS)},
  year      = {2017}
}

@misc{liao2025beyondmimicmotiontrackingversatile,
      title={BeyondMimic: From Motion Tracking to Versatile Humanoid Control via Guided Diffusion}, 
      author={Qiayuan Liao and Takara E. Truong and Xiaoyu Huang and Yuman Gao and Guy Tevet and Koushil Sreenath and C. Karen Liu},
      year={2025},
      eprint={2508.08241},
      archivePrefix={arXiv},
      primaryClass={cs.RO},
      url={https://arxiv.org/abs/2508.08241}, 
}

@InProceedings{peebles2023dit,
  author    = {Peebles, William and Xie, Saining},
  title     = {Scalable Diffusion Models with Transformers},
  booktitle = {Proceedings of the IEEE/CVF International Conference
               on Computer Vision (ICCV)},
  pages     = {4195--4205},
  year      = {2023}
}

@misc{yin2025unitrackerlearninguniversalwholebody,
      title={UniTracker: Learning Universal Whole-Body Motion Tracker for Humanoid Robots}, 
      author={Kangning Yin and Weishuai Zeng and Ke Fan and Minyue Dai and Zirui Wang and Qiang Zhang and Zheng Tian and Jingbo Wang and Jiangmiao Pang and Weinan Zhang},
      year={2025},
      eprint={2507.07356},
      archivePrefix={arXiv},
      primaryClass={cs.RO},
      url={https://arxiv.org/abs/2507.07356}, 
}

@InProceedings{lugmayr2022repaint,
  author    = {Lugmayr, Andreas and Danelljan, Martin and Romero, Andres
               and Yu, Fisher and Timofte, Radu and Van Gool, Luc},
  title     = {{RePaint}: Inpainting Using Denoising Diffusion
               Probabilistic Models},
  booktitle = {Proceedings of the IEEE/CVF Conference on Computer
               Vision and Pattern Recognition (CVPR)},
  pages     = {11461--11471},
  year      = {2022}
}

@misc{tevet2022humanmotiondiffusionmodel,
      title={Human Motion Diffusion Model}, 
      author={Guy Tevet and Sigal Raab and Brian Gordon and Yonatan Shafir and Daniel Cohen-Or and Amit H. Bermano},
      year={2022},
      eprint={2209.14916},
      archivePrefix={arXiv},
      primaryClass={cs.CV},
      url={https://arxiv.org/abs/2209.14916}, 
}

@misc{luo2024universalhumanoidmotionrepresentations,
      title={Universal Humanoid Motion Representations for Physics-Based Control}, 
      author={Zhengyi Luo and Jinkun Cao and Josh Merel and Alexander Winkler and Jing Huang and Kris Kitani and Weipeng Xu},
      year={2024},
      eprint={2310.04582},
      archivePrefix={arXiv},
      primaryClass={cs.CV},
      url={https://arxiv.org/abs/2310.04582}, 
}

@misc{luo2026sonicsupersizingmotiontracking,
      title={SONIC: Supersizing Motion Tracking for Natural Humanoid Whole-Body Control}, 
      author={Zhengyi Luo and Ye Yuan and Tingwu Wang and Chenran Li and Fernando Castañeda and Sirui Chen and Zi-Ang Cao and Jiefeng Li and David Minor and Qingwei Ben and Jinhyung Park and David Sami and Zi Wang and Xingye Da and Runyu Ding and Cyrus Hogg and Lina Song and Edy Lim and Eugene Jeong and Tairan He and Haoru Xue and Wenli Xiao and Simon Yuen and Jan Kautz and Yan Chang and Umar Iqbal and Linxi "Jim" Fan and Yuke Zhu},
      year={2026},
      eprint={2511.07820},
      archivePrefix={arXiv},
      primaryClass={cs.RO},
      url={https://arxiv.org/abs/2511.07820}, 
}

@conference{AMASS:ICCV:2019,
  title = {{AMASS}: Archive of Motion Capture as Surface Shapes},
  author = {Mahmood, Naureen and Ghorbani, Nima and Troje, Nikolaus F. and Pons-Moll, Gerard and Black, Michael J.},
  booktitle = {International Conference on Computer Vision},
  pages = {5442--5451},
  month = oct,
  year = {2019},
  month_numeric = {10}
}

@misc{ho2020denoisingdiffusionprobabilisticmodels,
      title={Denoising Diffusion Probabilistic Models}, 
      author={Jonathan Ho and Ajay Jain and Pieter Abbeel},
      year={2020},
      eprint={2006.11239},
      archivePrefix={arXiv},
      primaryClass={cs.LG},
      url={https://arxiv.org/abs/2006.11239}, 
}

@article{Kimodo2026,
  title={Kimodo: Scaling Controllable Human Motion Generation},
  author={Rempe, Davis and Petrovich, Mathis and Yuan, Ye and Zhang, Haotian and Peng, Xue Bin and Jiang, Yifeng and Wang, Tingwu and Iqbal, Umar and Minor, David and de Ruyter, Michael and Li, Jiefeng and Tessler, Chen and Lim, Edy and Jeong, Eugene and Wu, Sam and Hassani, Ehsan and Huang, Michael and Yu, Jin-Bey and Chung, Chaeyeon and Song, Lina and Dionne, Olivier and Kautz, Jan and Yuen, Simon and Fidler, Sanja},
  journal={arXiv:2603.15546},
  year={2026}
}

@article{JMLR:v21:20-212,
  author  = {Sanmit Narvekar and Bei Peng and Matteo Leonetti and Jivko Sinapov and Matthew E. Taylor and Peter Stone},
  title   = {Curriculum Learning for Reinforcement Learning Domains: A Framework and Survey},
  journal = {Journal of Machine Learning Research},
  year    = {2020},
  volume  = {21},
  number  = {181},
  pages   = {1--50},
  url     = {http://jmlr.org/papers/v21/20-212.html}
}

@misc{huang2025learninghumanoidstandingupcontrol,
      title={Learning Humanoid Standing-up Control across Diverse Postures}, 
      author={Tao Huang and Junli Ren and Huayi Wang and Zirui Wang and Qingwei Ben and Muning Wen and Xiao Chen and Jianan Li and Jiangmiao Pang},
      year={2025},
      eprint={2502.08378},
      archivePrefix={arXiv},
      primaryClass={cs.RO},
      url={https://arxiv.org/abs/2502.08378}, 
}

@article{Zakka_mjlab_A_Lightweight_2026,
    author = {Zakka, Kevin and Liao, Qiayuan and Yi, Brent and Le Lay, Louis and Sreenath, Koushil and Abbeel, Pieter},
    title = {{mjlab: A Lightweight Framework for GPU-Accelerated Robot Learning}},
    url = {https://arxiv.org/abs/2601.22074},
    year = {2026}
}

@misc{chen2024diffusionforcingnexttokenprediction,
      title={Diffusion Forcing: Next-token Prediction Meets Full-Sequence Diffusion}, 
      author={Boyuan Chen and Diego Marti Monso and Yilun Du and Max Simchowitz and Russ Tedrake and Vincent Sitzmann},
      year={2024},
      eprint={2407.01392},
      archivePrefix={arXiv},
      primaryClass={cs.LG},
      url={https://arxiv.org/abs/2407.01392}, 
}

@misc{ho2022classifierfreediffusionguidance,
      title={Classifier-Free Diffusion Guidance}, 
      author={Jonathan Ho and Tim Salimans},
      year={2022},
      eprint={2207.12598},
      archivePrefix={arXiv},
      primaryClass={cs.LG},
      url={https://arxiv.org/abs/2207.12598}, 
}

@article{lin2025simgenhoi,
  title     = {SimGenHOI: Physically Realistic Whole-Body Humanoid-Object Interaction via Generative Modeling and Reinforcement Learning},
  author    = {Lin, Yuhang and Xie, Yijia and Xie, Jiahong and Huang, Yuehao and Wang, Ruoyu and Lv, Jiajun and Ma, Yukai and Zuo, Xingxing},
  journal   = {arXiv preprint arXiv:2508.14120},
  year      = {2025}
}

@misc{tessler2024maskedmimicunifiedphysicsbasedcharacter,
      title={MaskedMimic: Unified Physics-Based Character Control Through Masked Motion Inpainting}, 
      author={Chen Tessler and Yunrong Guo and Ofir Nabati and Gal Chechik and Xue Bin Peng},
      year={2024},
      eprint={2409.14393},
      archivePrefix={arXiv},
      primaryClass={cs.AI},
      url={https://arxiv.org/abs/2409.14393}, 
}

@article{Huang_2025,
   title={Diffuse-CLoC: Guided Diffusion for Physics-based Character Look-ahead Control},
   volume={44},
   ISSN={1557-7368},
   url={http://dx.doi.org/10.1145/3731206},
   DOI={10.1145/3731206},
   number={4},
   journal={ACM Transactions on Graphics},
   publisher={Association for Computing Machinery (ACM)},
   author={Huang, Xiaoyu and Truong, Takara and Zhang, Yunbo and Yu, Fangzhou and Sleiman, Jean Pierre and Hodgins, Jessica and Sreenath, Koushil and Farshidian, Farbod},
   year={2025},
   month=July, pages={1–12} }

@misc{kalaria2025dreamcontrolhumaninspiredwholebodyhumanoid,
      title={DreamControl: Human-Inspired Whole-Body Humanoid Control for Scene Interaction via Guided Diffusion}, 
      author={Dvij Kalaria and Sudarshan S Harithas and Pushkal Katara and Sangkyung Kwak and Sarthak Bhagat and Shankar Sastry and Srinath Sridhar and Sai Vemprala and Ashish Kapoor and Jonathan Chung-Kuan Huang},
      year={2025},
      eprint={2509.14353},
      archivePrefix={arXiv},
      primaryClass={cs.RO},
      url={https://arxiv.org/abs/2509.14353}, 
}
%%%%%%%%%%%%%%%%%%%%%%%%%%%%%%%%%%%%%%%%%%%%%%%%%%%%%%%%%%%%%%%%%%%%%%%%%%%%%%%%
\appendix
\section{Appendix}
\label{sec:appendix}

To keep the exact settings reproducible and machine-readable, we provide all remaining hyperparameters as Python configuration files as supplementary material:
\begin{itemize}[noitemsep]
  \item the motion-generator backbone and training settings;
  \item the pipeline and curriculum hyperparameters;
  \item the actor--critic network architecture;
  \item the motion-tracker reward terms;
  \item the PPO optimisation hyperparameters;
  \item the motion-tracking policy observation terms;
  \item the domain randomisation and reference-state initialisation ranges.
\end{itemize}

We will open-source the codebase upon acceptance of the paper.
\end{document}